\pdfoutput=1
\documentclass{article}
\usepackage{preprintstyle,times}

\usepackage{hyperref}
\usepackage{url}
\usepackage{amsmath,amssymb}
\usepackage{graphicx}
\usepackage{booktabs}
\usepackage{multirow}
\usepackage{xcolor}

\newcommand{\nuc}{\nu}
\newcommand{\absh}{\nu\!\cdot\!n_T}
\newcommand{\best}[1]{\textbf{#1}}

\title{JEPA Learns What the Mask \\ Leaves Unrecoverable}

\author{Peng~Xie and Amr~Alanwar%
\thanks{Peng Xie and Amr Alanwar are with the TUM School of Computation, Information and Technology, Department of Computer Engineering, Technical University of Munich, 74076 Heilbronn, Germany (e-mail: p.xie@tum.de; alanwar@tum.de).}%
}
\hypersetup{pdftitle={JEPA Learns What the Mask Leaves Unrecoverable},pdfauthor={Peng Xie, Amr Alanwar}}

\begin{document}
\maketitle

\begin{abstract}
Joint-embedding predictive architectures are unusually sensitive to how the input is masked:
block masks work, scattered masks do not, and the explanations are empirical. We give a
measurement account. A mask is a linear measurement, and in a compactly supported wavelet basis
every atom whose support lies inside the hidden region falls in the measurement's null space and
leaves no trace in the data. The JEPA loss asks only that the encoded context suffice for the target, so a target that a
low-level prior can recover admits a shortcut, one the moving-average target encoder can make
self-consistent. What removes the shortcut is the coarse-scale content the mask leaves
unrecoverable, provided enough context stays within reach of each target. We score that content before training and test the account's
distinctive predictions in 151 pre-training runs. On ImageNet-100, strip masks match blocks in
area and contiguity yet are recoverable, and they land at 40.3\% linear top-1, beside random masks at
40.8\%, against 64.3\% for blocks; within one geometry family, the placements that leave the least
unrecoverable content lose 6.5 points to those that leave the most, over five seed pairs;
pixel targets span 7 points where latent targets span 25; and against a frozen target the gap
between random and block masks, 19 points on the same kind of GPU, closes to 1.5, so the geometry acts through the
target the encoder produces for itself. On UCF101 the masking ratio decides which condition, content or reach, binds;
removing whole frames, unrecoverable in space but recoverable from neighbouring frames, is
worst at both ratios; and on V-JEPA's own masks, batching them intact instead of truncated
changes little (36.0\% against 35.1\%), whereas making 100 target tokens inside the blocks
visible lifts them to 48.7\% and hiding 100 context tokens outside the blocks does not (33.7\%).
\end{abstract}

\section{Introduction}

I-JEPA predicts the representation of hidden image regions from visible ones, and its authors
report that the mask decides whether the features are semantic: four target blocks, each
15--20\% of the image, reach 54.2\% on low-shot ImageNet, random masking of the same area
17.6\% \citep{assran2023ijepa}. V-JEPA repeats the finding on video \citep{bardes2024vjepa}, A-JEPA
finds the recipe transfers badly to spectrograms \citep{fei2023ajepa}, and MAE uses random
masking at 75\% \citep{he2022mae}. I-JEPA states two principles, large enough
targets and an informative enough context, and SimMIM a distance heuristic
\citep{xie2022simmim}; neither says which quantity a sweep optimises, so a recipe tuned on one
modality is re-tuned on the next.

We give a measurement account of both principles. The encoder sees only the context patches, $y = P_C x$, so masking is
a linear measurement; in a compactly supported wavelet basis $\Psi$ its matrix is $A = P_C \Psi$,
and every atom whose support lies inside the unobserved region satisfies $A e_i = 0$:
$P_C(x + c\psi_i) = P_C x$ for every $c$, so the measurement alone does not determine $c$. The
null space is the same size for every mask of one budget; the geometry sets which atoms fall in
it, and the energy natural signals place on those atoms is what the measurement does not carry and what the encoder, not a shortcut, must supply
from content. We write $\nuc$ for that
fraction, our instrument.

Four controls test the account, each decoupling two otherwise confounded variables. Strips, contiguous
and block-sized but recoverable, should behave like scattered masks, and do. Whole frames,
unrecoverable in space but recoverable from neighbouring frames, should be the worst video mask,
and are. A pixel target, which the encoder cannot pull toward a shortcut, should care less about
geometry, and at our budget does. A frozen semantic target, which no low-level map predicts,
should make the mask almost irrelevant to the linear probe, and does. Enough unrecoverable content is the first condition; a second, enough visible context within
reach of each target, takes over when context is scarce, and it is why a hundred visible tokens
inside V-JEPA's hidden blocks are worth thirteen points.

\paragraph{Contributions.}
\begin{itemize}\itemsep1pt
\item We give a measurement account of what a mask controls: a hidden region annihilates an
atom of scale $s$ only if its inscribed square reaches $s$, so what a low-level prior can
recover is set by depth, not by area or contiguity; and because the JEPA objective
asks only for sufficiency, the encoder has no reason to represent what such a prior recovers
(Section~\ref{sec:method}). Two conditions follow, enough unrecoverable content and enough
reachable context (Section~\ref{sec:plateau}).
\item We test the account's distinctive predictions on ImageNet-100 with eight families at
matched budget and context: strips behave like scattered masks before and after training, 24
points below blocks; the linear probe rises and a pixel probe falls with unrecoverable content,
a change of level rather than of quality; selecting placements within one family moves the
probe by $6.5 \pm 2.3$ points over five seed pairs; pixel targets span 7 points where latent
targets span 25; and against a frozen target random and block masks land within 1.5 points of
each other, where the moving target separates them by 19 on the same kind of GPU (Sections~\ref{sec:stage1}--\ref{sec:mae}).
\item On UCF101 the same account, with temporal copying as a prior, orders four families
exactly at 50\% masking; at 90\% the share of deep targets turns from asset to liability;
whole-frame removal is worst at both ratios; and inside V-JEPA's own blocks a hundred visible
tokens are worth thirteen points while the public collator's truncation changes little (Section~\ref{sec:video}).
\item We score unrecoverable coarse content before training with $\nuc$ (Equation~\ref{eq:nu}).
Across our
22 configurations it ranks them as the inscribed square and SimMIM's AvgDist do, and two pools that share those two but differ in $\nuc$ end 7.4
points apart (Section~\ref{sec:plateau}, Appendix~\ref{app:avgdist}).
\end{itemize}

\section{Related work}

\paragraph{What masked models learn.}
\citet{zhang2022howmask}, \citet{kong2023occlusion}, \citet{kong2023hier}, \citet{xie2023dark} and
\citet{park2023what} analyse what masked image modelling learns and how the ratio sets it;
ColorMAE \citep{hinojosa2024colormae}, its video and audio extension \citep{bhowmik2025structured}
and masked frequency modelling \citep{xie2023mfm} shape the mask's spectrum by design, the
quantity that $\nuc$ reads off any mask; none treats the mask as a measurement, the null-space
view we take from compressed sensing \citep{donoho2006cs,candes2006stable}.
\citet{lee2021predicting}, \citet{tsai2021multiview} and \citet{liu2022identifiability} analyse
what masked prediction recovers for a fixed target; Section~\ref{sec:method} makes the argument
for a target the encoder produces itself. Block masks with a latent target predate I-JEPA, with a fixed tokenizer in BEiT
\citep{bao2022beit} and a moving average in data2vec, iBOT and MSN
\citep{baevski2022data2vec,zhou2022ibot,assran2022msn}. \citet{littwin2024jepa} and
\citet{balestriero2024reconstruction} argue latent and pixel targets learn different features,
MaskFeat \citep{wei2022maskfeat} and I-JEPA \citep[Table 7]{assran2023ijepa} compare targets at
one mask, and Section~\ref{sec:mae} measures the difference family by family. MAE loses 9.6
linear-probe points to block masking at 75\% \citep[Table 1f]{he2022mae} and data2vec 2.0 finds
block-shaped and random visible regions within a point under fine-tuning \citep[Table 6]{baevski2023data2vec2}.
\citet{pathak2016context} first saw mask geometry set the level of the features. On video, VideoMAE \citep{tong2022videomae} and ST-MAE
\citep{feichtenhofer2022mae_video} find whole-frame removal the worst mask for a pixel target and
attribute it to leakage from neighbouring frames, which motion-guided masks block
\citep{huang2023mgmae,fan2023mgm} and SiamMAE \citep{gupta2023siammae} exploits; Section~\ref{sec:video}
finds the same order under a latent target.

\paragraph{Mask distance and adaptive masking.} SimMIM \citep{xie2022simmim} defines AvgDist, the mean distance
from a masked pixel to the nearest visible one, and explains why accuracy peaks at a middle distance: too small a distance lets pixels be
copied, too large makes the task too hard. I-JEPA's ablations of target and
context scale \citep[Tables 8--9]{assran2023ijepa} show the same two failure sides. Both are
empirical proxies of recoverability, as is InfoMin's rule that two views should share
task-relevant information and no more \citep{tian2020infomin}; the account names the quantity
behind them, and the controls of Sections~\ref{sec:stage1} and~\ref{sec:mae} separate it from
area, contiguity and the objective. Across our 22 configurations AvgDist, the inscribed square and $\nuc$ rank alike
($\rho = +0.766$, $+0.768$, $+0.770$), as measures of one quantity should; the pool test of
Section~\ref{sec:plateau} shows $\absh$ also carries what the
inscribed square and AvgDist do not (Appendix~\ref{app:avgdist}). AttMask \citep{kakogeorgiou2022attmask}, SemMAE \citep{li2022semmae}, HPM
\citep{wang2023hpm}, AdaMAE \citep{bandara2023adamae} and ADIOS \citep{shi2022adios} select
masks with model feedback and push toward harder ones; within our four block families the least
recoverable by PSNR, quadrants, is the worst, because beyond that point context reachability decides (Section~\ref{sec:plateau}).

\section{Masking is a linear measurement}
\label{sec:method}

Prior analyses ask what the model learns; we ask which part of the target no method can recover
knowing only that the signal is piecewise smooth. That part is computable before training, and
it is what the learner must supply from content.

\subsection{The objective demands sufficiency and nothing else}
Let $t = \bar f(x)_T$ be the moving-average target encoder's output on the target patches,
$f$ the context encoder and $g$ the predictor. We write the loss as $\mathbb{E}\|g(f(x_C)) - t\|^2$. The recipes use smooth-$\ell_1$ on images and $\ell_1$ on video; there the conditional median or Huber location replaces the mean, and the Bayes risk replaces the variance below.
For fixed $f$ the optimal $g$ is the conditional mean, so the loss at that $g$ equals
$\mathbb{E}\,\mathrm{Var}[t \mid f(x_C)]$. Because $f(x_C)$ is a function of $x_C$,
\begin{equation}
\mathbb{E}\,\mathrm{Var}[t \mid f(x_C)] \;\ge\; \mathbb{E}\,\mathrm{Var}[t \mid x_C],
\label{eq:sufficiency}
\end{equation}
with equality exactly when $\mathbb{E}[t \mid f(x_C)] = \mathbb{E}[t \mid x_C]$ almost surely, that is, when $f$ is sufficient for the conditional mean. Training pushes $f$ toward
sufficiency and, within what the predictor can compute, imposes no other requirement. If a low-level map $\ell$, say local interpolation, already achieves
$\mathbb{E}\,\mathrm{Var}[t \mid \ell(x_C)] \approx \mathbb{E}\,\mathrm{Var}[t \mid x_C]$,
then $f = \ell$ is near-optimal and the encoder has no reason to represent content. Because
$t$ is produced by the encoder's own moving average, that solution can become self-consistent under such a mask: an encoder that takes the shortcut makes its own targets recoverable \citep[the self-distillation dynamics of][]{grill2020byol,tian2021dynamics}. A frozen target is the direct test: a target the encoder cannot move cannot be made recoverable. Section~\ref{sec:mae} finds that it closes the random-versus-block gap from 19 points to 1.5 on the same kind of GPU.

\subsection{The measurement operator and its null space}
The encoder observes $y = P_C x$ with $P_C$ a row-subsampled identity. In a compactly supported wavelet basis $\Psi$
\citep{daubechies1988wavelets,mallat1989multiresolution}, $y = P_C\Psi a = A a$. Let $\psi_i$ be an atom whose support
misses $C$. Then $A e_i = 0$ and
\begin{equation}
P_C\,(x + c\,\psi_i) \;=\; P_C x \qquad \text{for every } c \in \mathbb{R}.
\label{eq:null}
\end{equation}
The data constrains $c$ not at all, so whatever a reconstruction puts there comes from its
prior: an $\ell_1$ prior on an orthonormal basis sets it to zero, at a cost equal to the energy
the signal places on $\psi_i$. Total-variation inpainting, a prior that couples coefficients, does
no better than that oracle on real photographs (Appendix~\ref{app:realenergy}).

\begin{figure}[t]
\centering
\includegraphics[width=0.9\linewidth]{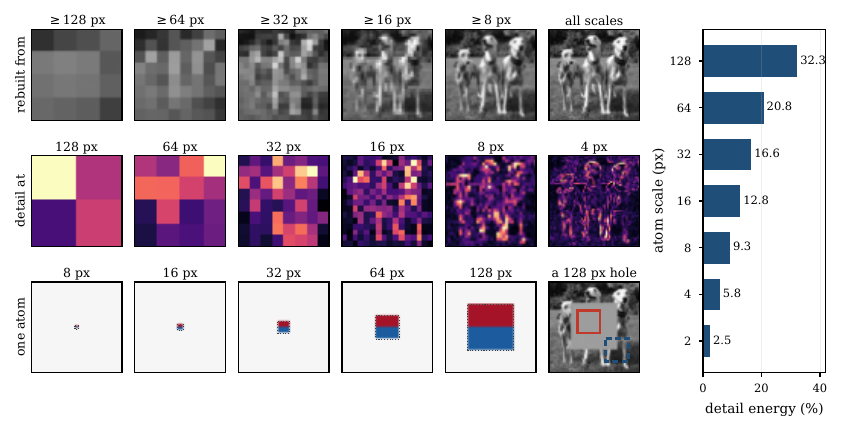}
\caption{\textbf{Top:} an image rebuilt from atoms of decreasing scale. \textbf{Middle:} where
its detail lives at each scale. \textbf{Bottom:} an atom of scale $s$ on its $s\times s$ support,
and two 64-pixel atoms against a 128-pixel hole: the red one lies inside it and is annihilated,
the blue one protrudes and is not. \textbf{Right:} detail energy by scale over 512 validation images.}
\label{fig:atoms}
\end{figure}

Whether an atom is annihilated depends on one geometric fact. A Haar atom of scale $s$ has an
$s\times s$ support, so a hole hides it only if its largest inscribed square reaches
$s$. Area, perimeter and contiguity do not enter. Natural images place most detail energy on
coarse atoms: measured on 512 ImageNet-100 images, the shares at 32, 64 and 128 pixels are
16.6, 20.8 and 32.3\% (Figure~\ref{fig:atoms}, Appendix~\ref{app:derivation}). We therefore score a mask by the energy-weighted fraction
of coarse atoms it annihilates,
\begin{equation}
\nuc(C,T) \;=\; \sum_{s \in \{32,64,128\}} w_s \cdot
\frac{\#\{\text{$s$-windows meeting } T \text{ with no context patch}\}}{\#\{\text{$s$-windows meeting } T\}},
\label{eq:nu}
\end{equation}
with $w_s$ the renormalised energy shares $(0.24, 0.30, 0.46)$. Scales up to one patch are hidden for
every mask at any budget and contribute a constant, so we drop them.
An $s$-window is an $s\times s$ square at any patch shift, so the score does not depend on where
the wavelet grid falls. Equation~\ref{eq:nu} reads only the mask, in a millisecond.
On video the same construction runs over space-time atoms indexed by a temporal and a spatial
extent, with the weights measured on real clips (Section~\ref{sec:video}).

\subsection{Fraction, amount, and reachability}
$\nuc$ is a fraction of the target's $n_T$ patches, the budget. What the encoder must supply is an
amount, so we also use $\absh$, the unrecoverable coarse content per sample; the two agree at
fixed budget only. A second quantity is needed because annihilating content is necessary but not
sufficient: targets far from every visible patch turn prediction into guessing. We measure that by reachability, the fraction of target patches with a context patch among their eight neighbours; on video we use the share of
targets three or more cells from any visible token (Section~\ref{sec:video}).

\section{Experimental setup}
\label{sec:setup}

\paragraph{Images.} We pre-train I-JEPA with ViT-S/16 \citep{dosovitskiy2021vit} on ImageNet-100 \citep{deng2009imagenet,tian2020cmc} (126{,}689 images, 100 classes) for 300 epochs at batch 256, using the official encoder, predictor, optimiser and
moving-average schedule \citep{assran2023ijepa}, replacing the mask collator and running the
predictor once over the union of the targets (Appendix~\ref{app:impl}). Every configuration
runs two seeds unless stated. We evaluate by linear probe and weighted $k$-NN ($k=20$) on
average-pooled target-encoder tokens (online-encoder tokens for the pixel-target models and the frozen-target students), and by a pixel probe: linear regression from each frozen
patch token to its $8\times8\times3$ downsampled content, reported as PSNR on validation. Every
run is probed at its final checkpoint under one protocol (Appendix~\ref{app:impl}). Eight families are drawn in matched sets, with the same image, the same
number of target patches, the same number of context patches and the same context block, so
that only the target geometry differs: F1 random scatter, F2 dispersed scatter, F3 twelve
small blocks, F4 four blocks (I-JEPA's blocks without overlap), F5 two blocks, F6 one block,
F7 horizontal strips one patch tall, F8 rasterised quadrants. The same image configuration scores up to 4 points differently on H100 and RTX 5090, so every
image comparison stays within one kind (Appendix~\ref{app:impl}).

\paragraph{Video.} We pre-train a V-JEPA-style model with ViT-S on UCF101 split 1
\citep{soomro2012ucf101}, 16 frames at stride 4 and $112\times112$ resolution, tubelet 2 and
patch 8, giving an $8\times14\times14$ grid of 1568 tokens, for 200 epochs at batch 32, with
the architecture and loss of \citet{bardes2024vjepa}. Eight spatio-temporal families
are drawn at 90\% masking and four of them at 50\%, and we evaluate by linear probe and $k$-NN on split 1 with
three clips per video; $\pm$ is the sample standard deviation over seeds throughout. Appendix~\ref{app:impl} gives all
hyperparameters; Appendix~\ref{app:audio} reports a third modality where the account does not hold.

\section{Images: what the mask leaves unrecoverable decides what is learned}

\subsection{Depth, not area, sets what a low-level prior can recover}
\label{sec:stage1}
\begin{figure}[t]
\centering
\includegraphics[width=0.9\linewidth]{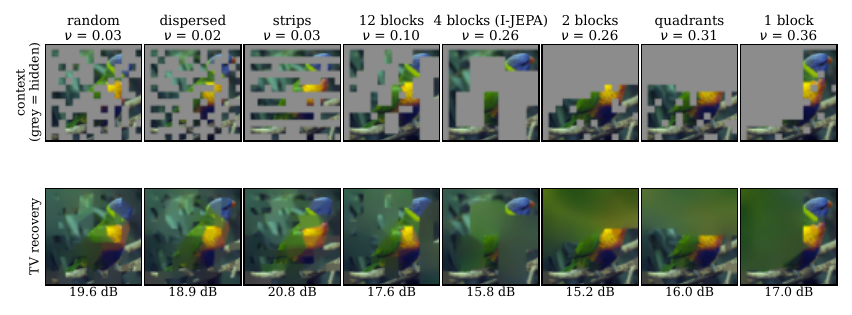}
\caption{Eight families on one image at a matched budget, from scattered to deep; grey is
everything the encoder does not see. Strips are as contiguous as a block and cover the same
area, yet on this image they are the \emph{most} recoverable family (per-image PSNR).}
\label{fig:inpaint}
\end{figure}

Three recovery algorithms rank eight families the same way (F4 and F5 within 0.04\,dB of each
other), by depth rather than by area or contiguity, and $\nuc$ tracks them at $\rho \le -0.93$ (Table~\ref{tab:stage1}: 40{,}000 paired
samples on 5{,}000 validation images, paired 95\% confidence interval $\pm0.06$\,dB against a
2--4\,dB gap between scattered and block families). Harmonic, total-variation
and wavelet-$\ell_1$ inpainting encode three notions of smoothness and agree on the ordering,
as the null-space argument requires. Strips are the discriminating case: contiguous, the same area as blocks, a longer perimeter,
but only one patch tall, and they land with the scattered families
(Figure~\ref{fig:inpaint}).

\begin{table}[t]
\centering
\caption{Depth decides recoverability; area and contiguity do not. Eight families at a
matched budget on 5{,}000 ImageNet-100 validation images, with the probe after 300 epochs of
pre-training on the right. Family-level $\rho(\nuc,\text{TV}) = -0.95$;
$\rho(\nuc,\text{linear}) = +0.71$.}
\label{tab:stage1}
\small
\begin{tabular}{llcccccc}
\toprule
 & Family & $\nuc$ & Insc.\ sq. & Harmonic & TV & Wavelet-$\ell_1$ & Linear top-1 \\
\midrule
F2 & dispersed          & 0.023 & 2.39 & 18.78 & \best{17.65} & 16.36 & 42.00 $\pm$0.51 \\
F1 & random             & 0.033 & 2.58 & 18.36 & 17.20 & 15.98 & 40.77 $\pm$1.99 \\
F7 & strips             & 0.027 & 2.44 & 18.25 & 17.03 & 15.73 & 40.29 $\pm$2.70 \\
F3 & 12 blocks          & 0.099 & 3.50 & 16.87 & 15.70 & 14.97 & 49.14 $\pm$1.90 \\
F4 & 4 blocks $\dagger$ & 0.264 & 6.39 & 14.87 & 13.77 & 13.94 & 64.30 $\pm$0.62 \\
F5 & 2 blocks           & 0.264 & 6.73 & 14.87 & 13.81 & 13.92 & \best{65.34} $\pm$0.08 \\
F8 & quadrants          & 0.305 & 7.00 & 14.19 & 13.06 & 13.45 & 56.95 $\pm$0.24 \\
F6 & 1 block            & 0.364 & 8.74 & 14.41 & 13.37 & 13.62 & 62.92 $\pm$0.03 \\
\bottomrule
\end{tabular}
\\[2pt]
{\footnotesize PSNR in dB on the target region, higher is more recoverable. $\dagger$ I-JEPA-like geometry. Recovery uses I-JEPA's
sampler, whose blocks may overlap (mean $n_T = 92.6$); training fixes $n_T = 92$ without overlap, where F4 has $\nuc = 0.18$ and the split is unchanged (Appendix~\ref{app:avgdist}).}
\end{table}

\subsection{What the prior cannot recover predicts what the encoder learns}
\label{sec:probe}
The four families below $\nuc = 0.1$ reach 43.1\% linear-probe top-1 and the four above it reach 62.4\%,
with no overlap (Table~\ref{tab:stage1}). The gap is 19.3 points, Mann--Whitney one-sided $p = 0.014$, the floor for four versus four; the boundary, drawn after the runs, sits in the widest gap of $\nuc$, from
0.099 to 0.264.
Strips sit at 40.3\%, next to random at 40.8\% and 24.0 points below four blocks.

\subsection{The effect is a change of level, not of quality}
\label{sec:levels}
The linear probe rises with $\nuc$ while the pixel probe falls, $\rho = +0.714$ against
$\rho = -0.905$ (exact one-sided $p = 0.002$). A merely harder task would move both the same way. Strips are the exception, pixel probe
18.9\,dB at 40\%: they lose pixel content without gaining a semantic level, which the account
does not explain. Intermediate checkpoints show the levels separating after epoch 100: the block families push
pixel content out, the scattered families do not move, and in linear probe the two groups never cross; easy masks
do not start fast, which counts against reading easy masks as a curriculum (Appendix~\ref{app:traj}).

\subsection{Selecting placements by unrecoverable content moves the representation}
\label{sec:causal}
Selecting masks by $\nuc$ inside one geometry family moves the linear probe by 6.5 points,
positive in all five seed pairs. For each image we draw 16 placements of four non-overlapping blocks and keep the one with the
highest or the lowest $\nuc$. Family, budget ($n_T = 92$ target and $n_C = 76$ context patches),
optimiser and schedule are identical; only the placement differs, and the inscribed square (4.74 against 5.61), AvgDist (1.52 against 1.93)
and reachability (0.58 against 0.41) move with it. The lowest of 16 has $\nuc = 0.151$, the
unselected draw 0.180 and the highest 0.230; on two H100 seeds they score $60.2 \pm 1.8$,
$64.3 \pm 0.6$ and $64.7 \pm 0.1$\%, on three RTX 5090 seeds $56.2 \pm 2.0$, $61.5 \pm 0.6$ and
$64.0 \pm 1.2$\%. Highest minus lowest at the same seed is $+3.1$ to $+9.0$ over the five
pairs, mean $6.5 \pm 2.3$, paired $t = 6.3$ (Appendix~\ref{app:avgdist}). The effect is asymmetric: the lowest selection loses 4.1 (H100) and 5.4 (RTX 5090) points to the unselected draw, the highest
gains 0.4 and 2.5.

\begin{figure}[t]
\centering
\includegraphics[width=0.9\linewidth]{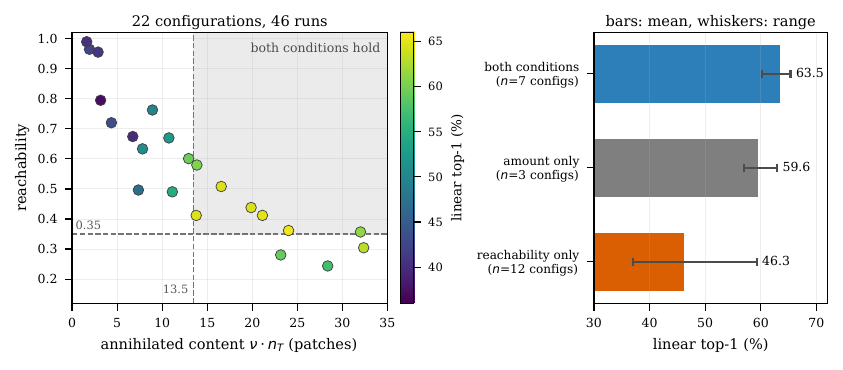}
\caption{The 22 configurations placed by annihilated content $\absh$ and reachability, coloured by linear-probe accuracy; dashed lines are the thresholds and the shaded quadrant is where both hold. Right: mean and range of each group.}
\label{fig:phase}
\end{figure}

\subsection{Two conditions and a plateau}
\label{sec:plateau}
Two inequalities computed from the mask separate our 22 configurations (Appendix~\ref{app:grids}) into a plateau and two
failure modes. Pooling the 22 configurations, 46 runs in all, each scored on the masks it
trained with, $\absh$ ranks them at $\rho = +0.84$, and the inscribed square and AvgDist
rank them alike (Appendix~\ref{app:avgdist}), as measures of one quantity should. A test whose
predictions and decision rule we fixed before training asks whether the amount of unrecoverable
content matters beyond the inscribed square and AvgDist: two pools of about 19{,}000
four-block placements each, built to share the inscribed square (5 patches), AvgDist (1.60
against 1.61) and the budget and to differ in $\nuc$ (0.154 against 0.197). The high-$\nuc$ pool
reaches $64.1 \pm 0.7$\% over three seeds, the low-$\nuc$ pool 56.7\% over the two seeds that
trained (its third collapsed). The pools also differ in
reachability (0.55 against 0.49), which moves with $\nuc$ at fixed budget (per-mask $\rho = -0.88$), so
the test separates the amount from the inscribed square and AvgDist, not from reachability (Appendix~\ref{app:avgdist}).
Figure~\ref{fig:phase} places every configuration on the two axes. With both thresholds read off the 46 sweep runs, the 7 of
22 configurations that satisfy both ($\absh \ge 13.5$, reach $\ge 0.35$), the plateau, average 63.5\% (range 60.2--65.3), the 3 that satisfy
only the first average 59.6\%, and the 12 that satisfy only the second average 46.3\%. Among the 22 sweep configurations the spread inside the plateau falls to 5 points from 28, and I-JEPA's block geometry at fixed budget sits inside it (Appendix~\ref{app:grids}). The thresholds summarise a graded response: the two test pools clear
them and still differ by 7.4 points.

\subsection{The effect runs through the target the encoder produces}
\label{sec:mae}
A pixel target removes nearly three quarters of the effect at this budget under linear probing;
a frozen latent target leaves 1.5 of the 19 points that separate random masks from four blocks on
the same kind of GPU (23.5 on the H100 runs of Table~\ref{tab:mae}). For the pixel target we keep the
encoder, predictor and masks, change the objective to per-patch normalised pixel reconstruction
with no moving average under MAE's learning-rate and weight-decay
settings, and probe
the online encoder (Appendix~\ref{app:impl}). The eight families span 6.8 points under the pixel
objective and 25.1 under I-JEPA, and the two curves cross between $\nuc = 0.10$ and $0.18$: masks
below the crossing gain 10--14 points from a pixel target, masks above it lose 1.5--6.2
(Table~\ref{tab:mae}); MAE's own recipe, random 75\%, scores 58.5\%, 1.3 points below the best
pixel-target family. For the frozen target we take the converged target encoder of a four-block run (61.4\%
linear on RTX 5090), never update it, and train the random and four-block families against it with
everything else unchanged, probing the online encoder (Appendix~\ref{app:frozen}). Random masks reach 67.3\% and four blocks 68.8\%, two seeds each, against 42.1\% and 61.2\% with
the moving target on the same hardware. We fixed the
rule before training: at most 8 points for the account, 15 or more against it.
Both students pass their teacher, as SALT finds for frozen teachers \citep{li2025salt}, and the pixel probe keeps
the direction of Section~\ref{sec:levels}: the geometry still sets how much low-level content the
encoder keeps, and the target sets the level it reaches (Appendix~\ref{app:frozen}). Mixing a quarter of random masks into block training scores 5.6 points below the interpolation
of the pure runs, so the low-level solution competes with the semantic one rather than seeding it
(Appendix~\ref{app:mixing}).

\begin{table}[t]
\centering
\caption{Mask geometry matters nearly four times more for a latent target than for a pixel target, and
the two curves cross. Same encoder, predictor and masks ($\ddagger$ excepted); the pixel objective uses MAE's optimiser
settings and no moving average (Appendix~\ref{app:impl}). Mean of two seeds.}
\label{tab:mae}
\small
\begin{tabular}{lccccc}
\toprule
Family & $\nuc$ & Pixel target & I-JEPA & Pixel $-$ I-JEPA & Pixel probe \\
\midrule
strips             & 0.02 & 52.93 $\pm$0.01 & 40.29 & $+12.64$ & 24.40 \\
dispersed          & 0.02 & 53.10 $\pm$0.40 & 42.00 & $+11.10$ & 23.90 \\
random             & 0.03 & 55.09 $\pm$0.30 & 40.77 & $+14.32$ & 23.73 \\
random 75\% $\ddagger$ & 0.08 & 58.48 $\pm$0.23 & --- & --- & 23.60 \\
12 blocks          & 0.10 & 59.19 $\pm$0.95 & 49.14 & $+10.05$ & 23.94 \\
4 blocks $\dagger$ & 0.18 & \best{59.76} $\pm$0.54 & 64.30 & $-4.54$ & 24.62 \\
2 blocks           & 0.26 & 59.11 $\pm$0.24 & \best{65.34} & $-6.23$ & 24.95 \\
quadrants          & 0.31 & 55.45 $\pm$0.44 & 56.95 & $-1.50$ & 25.12 \\
1 block            & 0.35 & 58.29 $\pm$0.83 & 62.92 & $-4.63$ & 24.36 \\
\midrule
\multicolumn{2}{l}{spread across families} & 6.8 & 25.1 & & \\
\bottomrule
\end{tabular}
\\[2pt]
{\footnotesize $\dagger$ I-JEPA-like geometry. $\ddagger$ MAE's own recipe. $\nuc$ at the training budget; pixel probe of the pixel-target models, in dB.}
\end{table}

\section{Video: the masking ratio decides which condition binds}
\label{sec:video}

Video gives a hole a second way to be shallow: a region deep in space can be restored by copying
the neighbouring frame, so the low-level priors the score is defined against must include
temporal copying. We measure the space-time detail energy of real clips over pairs of temporal
and spatial extents, find 57.5\% of it on atoms with temporal detail (Appendix~\ref{app:video3d}),
and define $\nuc^{3d}$ as Equation~\ref{eq:nu} taken over every pair of extents, fine and static ones included,
with that energy as weights. The
recovery baseline copies the nearest visible frame and total-variation inpaints what was never
visible. Eight families vary the two depths independently: block tubes are deep in both, frame
blocks only in space, whole frames deep in space with visible neighbours, scattered tokens in neither.

\begin{figure}[t]
\centering
\includegraphics[width=0.9\linewidth]{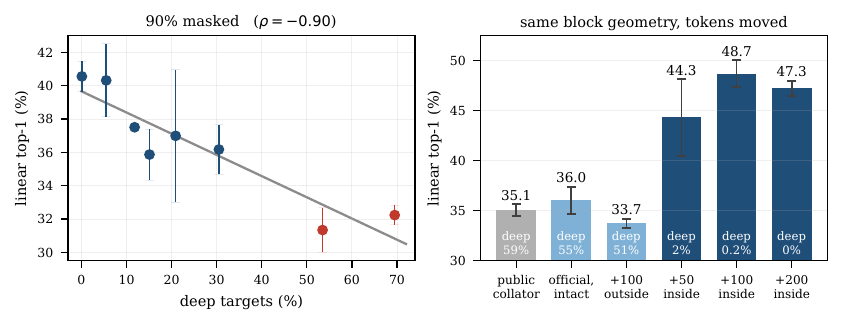}
\caption{Both $y$-axes start at 30. \textbf{Left:} at 90\% masking the families are ordered by the share of deep targets; red marks the two that remove whole time steps. \textbf{Right:} V-JEPA's block geometry intact, with 100 context tokens outside the blocks turned hidden, and with 50, 100 and 200 target tokens inside them made visible (grey: the public collator); bars carry the share of deep targets.}
\label{fig:video}
\end{figure}

\begin{table}[t]
\centering
\caption{The masking ratio decides which condition binds. UCF101 split 1, ViT-S, 200
epochs, mean of two seeds.}
\label{tab:video}
\small
\begin{tabular}{lcccc c cc}
\toprule
& \multicolumn{4}{c}{90\% masked ($n_C = 157$)} && \multicolumn{2}{c}{50\% masked ($n_C = 784$)} \\
\cmidrule(lr){2-5}\cmidrule(lr){7-8}
Family & $\nuc^{3d}$ & $R_{\text{low}}$ & Deep & Linear && $R_{\text{low}}$ & Linear \\
\midrule
random tokens    & 0.30 & 17.43 & 0.1\% & \best{40.56} $\pm$0.92 && 17.92 & 38.05 $\pm$2.07 \\
frame blocks      & 0.38 & 16.97 & 5.5\% & 40.33 $\pm$2.19 && 17.54 & 39.36 $\pm$0.22 \\
random tube       & 0.54 & 15.99 & 11.8\% & 37.51 $\pm$0.11 && --- & --- \\
tube, 1 block     & 0.52 & 16.45 & 20.9\% & 36.99 $\pm$3.98 && --- & --- \\
tube, 4 blocks    & 0.54 & 15.84 & 30.5\% & 36.17 $\pm$1.48 && 17.07 & \best{41.67} $\pm$2.22 \\
moving blocks     & 0.44 & 16.40 & 15.1\% & 35.87 $\pm$1.53 && --- & --- \\
half of time      & 0.44 & 16.52 & 69.5\% & 32.24 $\pm$0.58 && --- & --- \\
whole frames      & 0.39 & 17.16 & 53.5\% & 31.34 $\pm$1.33 && 18.91 & 33.66 $\pm$0.24 \\
\bottomrule
\end{tabular}
\\[2pt]
{\footnotesize $R_{\text{low}}$: PSNR in dB after temporal copying plus TV inpainting, higher
is more recoverable. Deep: share of targets three or more cells from any visible token. Spearman
$\rho$ with the linear probe, at 90\% then 50\%: $\nuc^{3d}$ $-0.24$, $+0.40$; $R_{\text{low}}$
$+0.14$, $\mathbf{-1.00}$; deep $\mathbf{-0.90}$, $+0.95$ ($-0.76$ and $+0.95$ under $k$-NN); the context
density of Table~\ref{tab:fix} gives $+0.69$ and $-1.00$. Without the random fill of the 90\% budget (Appendix~\ref{app:native}) tube and frame blocks score 35.6 $\pm$0.3 and
40.0 $\pm$1.2 over three RTX 5090 seeds.}
\end{table}

\subsection{At 50\% the first condition binds, at 90\% the second}
At 50\% masking, recoverability under the temporal-copy prior orders the four families
exactly, $\rho = -1.00$ ($\nuc^{3d}$ alone gives $+0.40$), and the deepest family wins; the share
of targets three or more cells from any visible token ranks them the same way except for a tie at
zero between the two worst ($\rho = +0.95$).
At 90\% that share's correlation changes sign, $\rho = -0.90$ over all eight families ($-0.76$ under $k$-NN;
Figure~\ref{fig:video}, left): every family
now hides plenty, since with only 157 context tokens even scattered masks reach $\nuc^{3d} = 0.30$, and what
limits the encoder is target depth.

\subsection{Whole-frame removal is the temporal analogue of strips}
Removing whole frames hides every patch of the frames it touches, the deepest possible hole
in space, yet it is worst at both ratios and the most recoverable of the four at 50\%
(18.91\,dB), because neighbouring frames restore it. It rules out spatial depth alone, as strips rule out area and contiguity. Its failure is not a probe artefact: the seed-0 gap to the other six families is 6.6 points under
the standard probe, 6.6 with its own context masks and 6.5 with one shared tubelet (Appendix~\ref{app:views}).

\subsection{Visible anchors inside the blocks, and a truncation in the public collator}
\label{sec:collator}
V-JEPA's published recipe draws two mask sets per step, eight short-range or two long-range blocks
extruded over all tubelets; the two sets hide 63\% and 81\% of a sample, not the 90\% usually
quoted \citep{bardes2024vjepa}. To batch variable-length masks, its public collator truncates every mask
to the batch minimum, which drops the last tubelets from context and targets alike
(Appendix~\ref{app:native}).

Table~\ref{tab:fix} compares the official masks batched intact with five other collators. Batching every
sample's official mask intact, with padding and an attention mask, scores 36.0\% against the
public collator's 35.1\%, inside the seed spread, and thinning the intact context at random to the truncated counts changes nothing (36.4\%, Appendix~\ref{app:native}). A hundred target tokens inside the blocks
turned visible lift the official masks to 48.7\% (three seeds, 47.2--49.5), a hundred context tokens
outside the blocks turned hidden do not (33.7\%), each a 100-token change to the intact budget. Fifty such anchors give 44.3\% and two hundred 47.3\%; at a hundred, one visible token per twelve
block cells, 0.2\% of the targets stay deep, and past that the anchors only remove unrecoverable content.

\begin{table}[t]
\centering
\caption{Where the visible tokens sit decides: V-JEPA's 8+2 block geometry as batched by six collators, UCF101 split 1, ViT-S, 200 epochs, seeds in parentheses.}
\label{tab:fix}
\footnotesize
\begin{tabular}{lcccc}
\toprule
V-JEPA's 8+2 masks, as batched & Density & Deep targets & Linear & $k$-NN \\
\midrule
public collator, batch-min truncation & 1.14 & 59\% & 35.06 $\pm$0.58 (2) & 26.74 \\
official masks intact, batched by padding & 1.76 & 55\% & 36.04 $\pm$1.36 (3) & 28.20 \\
\quad 100 context tokens outside the blocks hidden & 2.11 & 51\% & 33.73 $\pm$0.45 (3) & 27.41 \\
\quad 50 target tokens inside the blocks made visible & 2.71 & 2\% & 44.34 $\pm$3.83 (2) & 34.79 \\
\quad 100 inside & 3.65 & 0.2\% & \best{48.72} $\pm$1.35 (3) & \best{37.89} \\
\quad 200 inside & 5.55 & 0\% & 47.25 $\pm$0.73 (2) & 34.77 \\
\bottomrule
\end{tabular}
\\[2pt]
{\footnotesize Density: mean context tokens among the 26 space-time neighbours of a target; deep targets:
share of targets three or more cells from any visible token. Public-collator row trained on H100/H200, the others on RTX 5090 (the one configuration run on both, the count-fixing variant, agrees; Appendix~\ref{app:impl}); the count-fixing variant and its thinned and truncated rows are in Appendix~\ref{app:native}.}
\end{table}

That is the account's video instance: the same official blocks with a hundred tokens moved into
or out of the visible set land 15 points apart, and since thinning changes nothing, where the
visible tokens sit is what matters, not how many there are. With the public collator, V-JEPA's geometry sits below random
tubes on the same GPUs (35.1\% against 37.5\%), the reverse of V-JEPA's own ablation (Appendix~\ref{app:native}).

\section{Conclusion}

Masking is a linear measurement, and what it leaves unrecoverable to any low-level prior is
what pushes a joint-embedding predictive architecture to represent content instead of taking a
shortcut: strips and whole frames fail because a prior recovers them, blocks succeed because it
cannot. Good masks satisfy two conditions, enough unrecoverable content and enough reachable context,
and the masking ratio sets which one binds. The effect runs through the target the encoder
produces for itself: a pixel target shrinks it to a quarter and a frozen target to 1.5 points, so
the mask's job is to leave the encoder no low-level target it can make self-consistent. On
V-JEPA's own masks the collator's truncation changes little and a hundred visible anchors inside
the hidden blocks are worth thirteen: where the visible tokens sit decides what the encoder
learns, how many barely matters.

\textbf{Limitations.} The evidence rests on ViT-S, one dataset per modality and one to five seeds per
configuration, two for most. Both thresholds are read off the configurations
they separate; reachability rests on three amount-only configurations (Figure~\ref{fig:phase});
the pool test separates $\nuc$ from the inscribed square and AvgDist but not from
reachability. The frozen-target control has one four-block teacher and two
seeds per family; whether a low-level teacher would restore the gap is open (Appendix~\ref{app:frozen}).
On spectrograms the account fails (Appendix~\ref{app:audio}); Appendix~\ref{app:scope} states what
each result rests on.

\subsection*{Reproducibility statement}
Section~\ref{sec:method} states the assumptions behind $\nuc$ and Appendix~\ref{app:closed}
its closed form for random masks; the protocol is in Section~\ref{sec:setup} and every
hyperparameter, mask sampler and probe setting in Appendix~\ref{app:impl}. All 151 image and video pre-training runs and the 8 audio runs use published architectures on H100, H200 and RTX 5090 GPUs; relative to the official I-JEPA and V-JEPA recipes, and outside the pixel-target ablation and the frozen-target control, three things change: the mask collator, the key-padding attention mask that the padded video collators need, and a single predictor pass over the union of the targets; RTX 5090 video runs accumulate two batches of 16 (Appendix~\ref{app:impl}). Source code for the mask families, the score, the recovery baselines, the collators and the probes, with the per-run evaluation files behind every table, will be made publicly available.

\subsection*{Ethics statement}
The work studies pre-training objectives on public research datasets used under their terms of
access (Appendix~\ref{app:impl}); it involves no human subjects and no personal data beyond what
those datasets contain, and the generated images of Appendix~\ref{app:synthetic} are used under their
model's licence (Appendix~\ref{app:impl}).

\subsection*{AI use statement}
In this work, we used generative AI tools for tasks with required disclosure: an AI coding
assistant helped develop the measurement framing, formulated its mathematical claims and the short
derivations of Section~\ref{sec:method} and Appendix~\ref{app:derivation}, proposed and refined hypotheses
and control conditions, designed experiments, suggested their parameters and the pre-registered
decision rules, and gave feedback on them, implemented
the mask families, collators, training and evaluation code, reformatted the ImageNet-100 mirror
into image folders, and drafted interpretations of results. The synthetic images of
Appendix~\ref{app:synthetic} were generated with Stable Diffusion 1.4 as an object of study.
The work contains no formal theorems, no qualitative or thematic data analysis and no
translation, so those tasks are not applicable. We also used generative AI tools for tasks with recommended disclosure:
plotting code, cluster and cloud job scripts, drafting and editing the text, suggesting its
structure, brainstorming, formatting references, suggesting title options and keywords, and
searching for and summarising related work. The authors posed the initial question, chose
which directions, experiments and claims to pursue, and reviewed all AI-assisted work: every
number we report from our own runs is traced to the evaluation file of a logged run, the collator behaviour of
Section~\ref{sec:collator} was verified by direct measurement of the mask tensors, and the
citations were checked against their sources. We take responsibility for the final content of
this work, including text, claims or artifacts produced with the aid of generative AI.

\subsection*{Acknowledgements}
Experiments presented in this work were carried out using the CIT-TUM-HN cluster at TUM Campus Heilbronn.

\bibliography{references}
\bibliographystyle{preprintstyle}

\clearpage
\appendix
\section{Derivation of $\nuc$ and its weights}
\label{app:derivation}

\paragraph{Wavelet decomposition.} One level of the 2-D Haar transform replaces every
$2\times2$ cell by one average and three details. Applying it recursively to the averages
gives, for a $256\times256$ image, $3\cdot128^2$ atoms of support $2\times2$,
$3\cdot64^2$ of support $4\times4$, and so on to $3\cdot2^2$ of support $128\times128$,
plus four remaining averages: $49152 + 12288 + 3072 + 768 + 192 + 48 + 12 + 4 = 65536$
coefficients for 65{,}536 pixels. The transform is orthonormal, so detail energy plus
average energy equals image energy. On the $224\times224$ images we use, the seven-level
transform runs with periodic boundary handling, so the 128-pixel level is defined on the whole image.

\paragraph{Energy shares.} Measured over 512 ImageNet-100 validation images at $224\times224$,
the share of detail energy by atom scale is: 2\,px 2.5\%, 4\,px 5.8\%, 8\,px 9.3\%,
16\,px 12.8\%, 32\,px 16.6\%, 64\,px 20.8\%, 128\,px 32.3\%. The final approximation band
is excluded: with seven periodised levels on 224 pixels it holds four coefficients on wrapped
$128\times128$ supports, which we treat as the image mean rather than as atoms a hole can hide.

\paragraph{Which scales enter.} Atoms of 16 pixels and below fit inside a single patch, and every
target patch is unobserved, so those atoms are annihilated for every family at a matched
budget and contribute a constant. Equation~\ref{eq:nu} therefore uses 32, 64 and 128\,px
with the renormalised weights $(0.24, 0.30, 0.46)$.

\paragraph{A closed form for random masks.}\label{app:closed} For an i.i.d.\ mask with
unobserved fraction $p$ whose context is the complement of the target, an $m\times m$ window is
fully unobserved with probability $p^{m^2}$ and meets the target with probability $1-(1-p)^{m^2}$,
so $\nuc = \sum_s w_s\, p^{m_s^2}/\bigl(1-(1-p)^{m_s^2}\bigr)$ with $m_s = s/16$; for $p \ge 0.5$ the
denominator moves the third decimal only, and we use $\nuc \approx \sum_s w_s\, p^{(s/16)^2}$. At a fixed budget of 92 targets the context
averages 78 patches in our training-free statistics (training fixes 76), so $p = 0.60$ and the
window probabilities $p^{m^2}$ are $0.130$, $2.9\!\times\!10^{-4}$ and $\approx 0$ for $m = 2, 4, 8$, giving $\nuc \approx 0.031$
against $0.029$ measured for random scatter at that budget; Table~\ref{tab:stage1}, at I-JEPA's
variable sampler budget, gives $0.033$. At budgets 40, 60 and 120, with training-free contexts of 129, 109 and 52 patches, the formula
gives $0.003$, $0.009$ and $0.072$ against measured $0.005$, $0.010$ and $0.066$. The agreement is
approximate because our context is a contiguous block rather than an i.i.d.\ draw; on a
genuinely i.i.d.\ mask the formula matches simulation to within $0.001$. Applied to MAE-style
masking, where the context is exactly the complement and the assumption holds, it gives
$\nuc = 0.015$ at 50\% masking, $0.078$ at 75\% and $0.212$ at 90\%.

\section{Null-space energy on unmodified photographs}
\label{app:realenergy}

We decompose 96 unmodified validation images, mark every atom whose support lies entirely in
the unobserved region, and measure the share of detail energy those atoms carry. We also
compute an oracle: set the annihilated coefficients to zero and keep every other coefficient
at its true value, the error of any method that sets every annihilated coefficient to zero and
keeps every other coefficient exact; a prior that couples coefficients could in principle do better,
and on these images TV does not. Masks use a $16\times16$ grid with $n_T = 120$ of 256 patches (47\%).

\begin{center}
\small
\begin{tabular}{lccc}
\toprule
Family & Null-space energy & Oracle ceiling (dB) & TV inpainting (dB) \\
\midrule
strips     & 13.6\% & 19.77 & 18.47 \\
dispersed  & 14.1\% & 19.55 & 18.98 \\
random     & 15.0\% & 19.37 & 18.57 \\
12 blocks  & 19.0\% & 18.29 & 16.21 \\
4 blocks   & 23.6\% & 17.32 & 15.10 \\
2 blocks   & 27.7\% & 16.57 & 13.71 \\
1 block    & 29.5\% & 16.29 & 13.56 \\
quadrants  & 35.5\% & 15.21 & 12.90 \\
\bottomrule
\end{tabular}
\end{center}

Under scattered masks TV inpainting comes within 0.6--1.3\,dB of the oracle ceiling, so most
of its error is null space. Under block masks the ceiling itself drops by 2--4\,dB and TV falls
a further 2--3\,dB, because atoms that only partly protrude from a deep hole are weakly
constrained as well. A floor of 13--15\% is shared by every family: atoms no larger than one patch
are hidden whatever the geometry.

\section{The feature-space view, and a measure that fails}
\label{app:feature}

To ask what is predictable in the space the loss actually operates in, we freeze DINO \citep{caron2021dino}
ViT-B/16 and compare three predictors of target tokens from context tokens: the nearest context token,
inverse-distance weighting of the four nearest, and a 4-layer transformer trained on 4{,}488
images. Cosine similarity on 512 held-out images.

\begin{center}
\small
\begin{tabular}{lcccc}
\toprule
Family & Nearest ctx.\ token & Inverse distance & Trained predictor & Gap \\
\midrule
dispersed  & 0.597 & 0.695 & 0.678 & $-0.018$ \\
random     & 0.591 & 0.679 & 0.668 & $-0.012$ \\
strips     & 0.578 & 0.664 & 0.671 & $+0.007$ \\
12 blocks  & 0.552 & 0.617 & 0.631 & $+0.014$ \\
4 blocks   & 0.458 & 0.509 & 0.567 & $+0.058$ \\
2 blocks   & 0.458 & 0.508 & 0.565 & $+0.057$ \\
1 block    & 0.429 & 0.479 & 0.553 & $+0.074$ \\
quadrants  & 0.409 & 0.460 & 0.560 & $+0.100$ \\
\bottomrule
\end{tabular}
\end{center}

Under scattered masks a trained predictor does not beat inverse-distance weighting: the
shortcut exists in feature space as well as in pixels. Under the four large-block masks learning
is worth 0.06--0.10, and under 12 blocks 0.014; gaps are computed before rounding.

The gap does not order the block families: quadrants has the largest gap and the worst probe.
A frozen encoder that already has semantics finds large holes easy, so it cannot estimate
what is reachable for a learner starting from scratch. We use reachability, a geometric
quantity, instead.

Feeding TV-inpainted images through the same frozen encoder scores 0.28--0.42, below the
0.42--0.50 of simply averaging all context tokens. Blurred completions are off-manifold, so
the low-level baseline in feature space has to be interpolation and not pixel inpainting.

\section{$\nuc$ against AvgDist}
\label{app:avgdist}

Over the 22 configurations, ranking by Spearman correlation with linear-probe accuracy:

\begin{center}
\small
\begin{tabular}{lcc}
\toprule
Quantity & $\rho$ & $\rho$ after multiplying by $n_T$ \\
\midrule
AvgDist \citep{xie2022simmim} & $+0.766$ & $+0.802$ \\
Inscribed square              & $+0.768$ & $+0.846$ \\
$\nuc$                        & $+0.770$ & $+0.843$ \\
\bottomrule
\end{tabular}
\end{center}

Table~\ref{tab:stage1}'s $\nuc$ column uses I-JEPA's sampler, whose four blocks may overlap (mean
$n_T = 92.6$, F4 at $\nuc = 0.264$); training fixes $n_T = 92$ without overlap (F4 at $\nuc = 0.18$).
F4 and F5 tie at three decimals in that column (0.2641 against 0.2637); the correlations use the
unrounded values.
With the training values the family-level $\rho(\nuc, \text{linear})$ is $+0.79$ against $+0.71$ and the
four-versus-four split is unchanged; the recovery algorithms were run on the sampler's masks only. Multiplying $\nuc$ by $n_T$ helps, though not significantly: over 5{,}000
bootstrap resamples of the 22 configurations, $\rho(\absh) - \rho(\nuc)$ is $+0.07$ with 95\% interval
$[-0.02, +0.22]$. As rankers the three amounts are equivalent: $\rho(\absh) - \rho(\text{inscribed}\cdot n_T)$
is $0.00$ $[-0.08, +0.08]$ and $\rho(\absh) - \rho(\text{AvgDist}\cdot n_T)$ is $+0.04$ $[-0.07, +0.17]$, and
under $k$-NN $\absh$, the inscribed square times $n_T$ and AvgDist times $n_T$ rank at $+0.844$,
$+0.848$ and $+0.810$. Multiplying by $n_T$ raises all three point estimates, and the clearest
case for the amount over the fraction is one deep block over 40 target patches against four blocks
over 92: $\nuc$ 0.184 against 0.180 and AvgDist 1.71 against 1.66, yet 46.6\% against 64.3\%,
because $\absh$ is 7.4 against 16.6.

\paragraph{A test fixed before training.} The sweep never separates the three quantities, so we built
masks that do. From the four-block family at $n_T = 92$, $n_C = 76$ we drew candidate masks
exactly as training draws them, context thinning included, kept those whose largest inscribed
square is 5 patches and whose AvgDist lies in the densest 0.1-wide band, and split the survivors
by $\nuc$ into a low pool (19{,}194 masks, $\nuc = 0.154 \pm 0.005$, $\absh = 14.2$, AvgDist
1.596, reachability 0.553) and a high pool (19{,}177 masks, $\nuc = 0.197 \pm 0.008$, $\absh =
18.2$, AvgDist 1.608, reachability 0.492). Training draws each batch's masks from the pool
verbatim. Before running we wrote down the predictions: the null-space account predicts a gap
of three to four points in favour of the high pool, an account based on inscribed square or
AvgDist predicts none, and the decision rule was a gap of at least 2.5 points with
non-overlapping seeds. The high pool scored 64.16, 64.72 and 63.40\% ($k$-NN 55.8, 57.7,
55.2); the low pool 58.82 and 54.64\% ($k$-NN 49.7, 44.9), and its third seed collapsed at the
sixth epoch (effective rank of the target features below 20 from then on; linear 10.1\%).
Over the seeds that trained the gap is 7.4 points, larger than predicted, with no overlap. The
low pool is the more reachable of the two and both clear the threshold of Section~\ref{sec:plateau},
so a benefit of reachability, which is how the account uses it, runs against the observed
direction, while the amount of annihilated coarse content runs with it. The pools were not
matched on reachability (0.553 against 0.492), and over 1{,}600 masks drawn from them the per-mask
Spearman correlation between reachability and $\nuc$ is $-0.88$, so at this budget the two are
largely one variable: a target patch that borders the context lies in fewer fully hidden windows.
The test therefore separates $\nuc$ from the inscribed square and from AvgDist, but it cannot tell
a gain from more annihilated content apart from a gain from less reachability past the threshold;
a pool matched on reachability as well is the natural next test; we have built one from 400{,}000 candidates ($\nuc$ 0.163 against 0.188 at reachability
0.523 against 0.519, AvgDist 1.62 against 1.60 and equal context density). At this budget no four-block
mask hides a 128-pixel window, so the pools differ in the 32- and 64-pixel terms of $\nuc$ only.

\paragraph{Five seed pairs of the selection experiment.} Section~\ref{sec:causal} selects the
lowest or the highest of 16 placements. Seeds 0 and 1 ran on H100 (lowest 61.52 / 58.92,
highest 64.64 / 64.76); seeds 2--4 ran on RTX 5090 (lowest 54.38 / 55.74 / 58.34, highest
62.68 / 64.74 / 64.68). The five paired differences, highest minus lowest at the same seed on
the same hardware, are $+3.12$, $+5.84$, $+8.30$, $+9.00$ and $+6.34$: mean 6.52, SD 2.31,
paired $t = 6.31$ on four degrees of freedom. The unselected draw on the same hardware moved
from 64.30 (H100, two seeds) to 61.54 (RTX 5090, three seeds); the highest selection barely
moved (64.70 to 64.03), the lowest moved most (60.22 to 56.15). Against the unselected draw on
the same hardware the highest selection therefore gains 0.4 on H100 and 2.5 on RTX 5090 (per
seed $+1.2$, $+2.5$, $+3.7$) and the lowest loses 4.1 and 5.4: the loss below the unselected
draw is the larger half on both kinds of hardware, and the gain above it is resolved only on
the RTX 5090 seeds.

\section{Budget, block count and context}
\label{app:grids}

All entries use four non-overlapping blocks unless stated, two seeds, and masks scored with
the collator that trained them.

\begin{center}
\small
\begin{tabular}{llccccc}
\toprule
Group & Configuration & $n_T$ & $n_C$ & $\nuc$ & $\absh$ & Linear top-1 \\
\midrule
\multirow{5}{*}{budget}
 & 4 blocks & 40  & 128 & 0.079 & 3.2  & 36.96 \\
 & 4 blocks & 40  & 76  & 0.109 & 4.4  & 43.24 \\
 & 4 blocks & 60  & 108 & 0.112 & 6.7  & 39.77 \\
 & 4 blocks & 60  & 76  & 0.130 & 7.8  & 50.73 \\
 & 4 blocks & 120 & 48  & 0.267 & 32.0 & 60.98 \\
\midrule
\multirow{4}{*}{block count}
 & 3 blocks  & 92 & 76 & 0.216 & 19.9 & 64.34 \\
 & 6 blocks  & 92 & 76 & 0.141 & 12.9 & 59.25 \\
 & 8 blocks  & 92 & 76 & 0.117 & 10.7 & 52.23 \\
 & 12 blocks & 92 & 76 & 0.097 & 8.9  & 49.14 \\
\midrule
\multirow{3}{*}{depth vs.\ supervision}
 & 1 block  & 40 & 128 & 0.184 & 7.4  & 46.62 \\
 & 2 blocks & 60 & 108 & 0.185 & 11.1 & 54.40 \\
 & 1 block  & 60 & 108 & 0.230 & 13.8 & \best{64.43} \\
\bottomrule
\end{tabular}
\end{center}

The same account is consistent with the ablation in which \citet{assran2023ijepa} predict one
target block of the usual scale instead of four, which performs far below four blocks: at roughly 30 target tokens a
single block has $\absh$ near 5.5, well under the threshold.

Two entries are worth reading twice. At 40 targets, cutting the context from 128 to 76 raises
the probe by 6.3 points, because patches that are neither context nor target are also
unobserved and deepen the holes: $\nuc$ rises from 0.079 to 0.109. And at 60 predicted tokens
with 108 context patches, one deep block reaches 64.43, matching the 92-token reference with
fewer predictions and more visible input.

The block-count sweep at a fixed budget of 92 gives 62.92, 65.34, 64.34, 64.30, 59.25, 52.23,
49.14 for $k = 1, 2, 3, 4, 6, 8, 12$: an inverted U with a plateau at 2--4. Small $k$ fails
reachability, large $k$ fails the amount constraint.

\section{Batch-minimum truncation in the public collators}
\label{app:native}

Both public collators stack a batch of variable-length masks into one tensor by truncating
every mask to the batch minimum, keeping the lowest token indices. Two things follow: the
context shrinks as the batch grows, and because token indices are raster-ordered on images and
time-major on video, the tokens that are dropped always come from the same place, the bottom
rows of an image and the end of a clip.

The truncation is the slice \texttt{cm[:min\_keep]} applied to every mask in the
\texttt{\_\_call\_\_} method of the mask collator (of its \texttt{\_MaskGenerator} for V-JEPA): lines 164 and 167 of
\nolinkurl{src/masks/multiblock.py} in \nolinkurl{facebookresearch/ijepa} at commit
\texttt{52c1ae9}, and lines 197 and 200 of \nolinkurl{src/masks/multiblock3d.py} in
\nolinkurl{facebookresearch/jepa} at commit \texttt{51c59d5}.

\paragraph{Images.} Sampling I-JEPA's context block per image gives a mean of 79.8 patches;
after truncation the batch receives 62.6 at batch 8, 53.3 at batch 64 and 48.3 at batch 256.
Run under our protocol at batch 256, the native sampler scores 58.23\% with four predictor
passes and 59.55\% with one, against 64.30\% for four non-overlapping blocks at a fixed
budget. The native sampler and F4 differ in overlap, budget variance and context size at once, so we do not
attribute their difference to geometry alone, and in Figure~\ref{fig:phase} the native sampler
enters with the statistics of its own masks; F4 fixes the budget but also removes the overlap, and the clean image control is the same official geometry run
through a per-sample fixed-budget collator, which we leave to future work. Every
configuration in the main text except this native-sampler row is trained through the
fixed-budget path, while Table~\ref{tab:stage1}'s training-free columns use I-JEPA's sampler
(Appendix~\ref{app:avgdist}): the logs of the eight-family sweep and the $k$ grid show exactly 76
context and 92 target patches per image, and the budget grid varies both counts deliberately.

\paragraph{I-JEPA's published masking ablation.} I-JEPA's Table~6 reports, for ViT-B/16 at 1\%
low-shot on ImageNet-1k, 54.2\% for multi-block masking against 20.2\% for a single block of
scale 0.6, 15.5\% for rasterised quadrants and 17.6\% for random patches. By our thresholds the
single block and the rasterised layout fail the reachability condition and the random layout
fails the amount condition, the directions the account predicts; our amount-only group sits
only four points below the plateau, so the size of that failure at I-JEPA's scale and protocol
is not reproduced here.

\paragraph{Video.} Measured over 60 batches of 32 on our $8\times14\times14$ grid, V-JEPA's
short-range mask set has a per-sample mean of 579 context and 989 target tokens and its
long-range set 293 and 1275. After truncation the batch receives 349 and 102 context tokens.
The last two tubelets, which hold 25\% of the context before truncation, hold 3.1\% and 2.3\%
after it. The per-sample masking ratios are 63\% and 81\%; after truncation 78\% and 93.5\% of the clip is
hidden, so the 90\% figure usually quoted for the recipe comes from the batching, not from the
geometry. The symptom, masks that
cover only the first frames and index lists whose lengths do not add up, was reported in the
repository's issue tracker in March 2024 (facebookresearch/jepa, issue 50) without a cause; the
collator released with V-JEPA~2 \citep{assran2025vjepa2} keeps the same batch-minimum
truncation (\texttt{multiseq\_multiblock3d.py}).

\paragraph{Batching the official masks intact.} The control the truncation calls for keeps
each sample's official mask, pads every context and target index list to the batch maximum,
and passes a boolean validity mask through the encoder and the predictor as a key-padding mask
in attention, with the loss averaged over valid targets only. Nothing about the masks changes.
Batched this way the official geometry scores 36.04\% (three seeds: 34.84, 37.51, 35.77)
against the public collator's 35.06\% (35.47, 34.66, trained on H100/H200), a difference inside
the seed spread (Welch $t = 1.1$); thinning its context at random to the truncated counts, 349
and 102, gives 36.44\% (36.88, 36.00). The tokens the truncation removes are the
last tubelets of regions that remain visible in the earlier tubelets, and under the temporal
prior their marginal information is close to zero.

\paragraph{What the count-fixing variant does.} Our first variant (family
\texttt{vjepa\_geom}) draws the official blocks with the official generator, then fixes each
sample's target count at 989 (short-range) or 1275 (long-range): where the block union
overshoots the count, randomly chosen cells inside the union become context, and where it falls
short, randomly chosen cells outside become targets. Simulated on 400 samples per set, the two
cases are equally frequent; the overshoot half moves 86 (short-range) and 112 (long-range)
cells into the blocks, which cuts the share of long-range targets three or more cells from any
visible token from 71\% to 1.5\%, while the undershoot half leaves the blocks as they were. The
variant therefore trains on anchored blocks half of the time, which is why it lands between the
intact geometry (36.0\%) and the anchored one (48.7\%). Re-run on RTX 5090 with the same
batching as the new rows it scores 42.48\%, against 42.76 $\pm$ 1.29 on H100.

\paragraph{Anchors in, targets out.} The two collator families that separate the effect start from
the intact official masks and, in every sample and both sets, move $k$ randomly chosen cells:
family \texttt{vjepa\_pad\_in} turns $k$ target cells inside the block union into context,
\texttt{vjepa\_pad\_out} turns $k$ context cells outside the union into targets. Both were
checked cell by cell against the intact masks before training. Table~\ref{tab:reach} gives the
mask statistics behind Table~\ref{tab:fix}, measured over 192 samples per set.

\begin{table}[h]
\centering
\caption{Mask statistics of the collator variants: density (mean context tokens among the 26
space-time neighbours of a target) averaged over the two sets, and, over all targets of both sets, the share with a visible token
among their 26 neighbours, the share three or more cells from any visible token, and the mean
chessboard distance in cells from a target to the nearest visible token (AvgDist). Rows marked
$\ast$ were trained on H100/H200, the others on RTX 5090.}
\label{tab:reach}
\small
\begin{tabular}{lccccc}
\toprule
Masks, as batched & Density & Anchored & Deep & AvgDist & Linear \\
\midrule
public collator (batch-min truncation) $\ast$ & 1.14 & 17\% & 59\% & 3.83 & 35.1 \\
official masks intact & 1.76 & 23\% & 55\% & 3.52 & 36.0 \\
\quad context thinned to 349 / 102 & 0.92 & 22\% & 55\% & 3.53 & 36.4 \\
\quad 100 context tokens outside hidden & 2.11 & 28\% & 51\% & 3.39 & 33.7 \\
\quad 50 target tokens inside visible & 2.71 & 71\% & 2\% & 1.31 & 44.3 \\
\quad 100 inside visible & 3.65 & 88\% & 0.2\% & 1.12 & 48.7 \\
\quad 200 inside visible & 5.55 & 98\% & 0\% & 1.02 & 47.3 \\
count fixed by random flips $\ast$ & 2.89 & 55\% & 24\% & 2.09 & 42.8 \\
\quad context thinned to 349 / 102 $\ast$ & 1.43 & 44\% & 29\% & 2.29 & 36.9 \\
\quad batch-min truncation & 1.48 & 32\% & 46\% & 2.73 & 37.8 \\
\bottomrule
\end{tabular}
\end{table}

Density does not order the rows: hiding 100 tokens outside the blocks raises it from 1.76 to
2.11 while the probe falls, and thinning the intact context to 349 / 102 halves it while the
probe holds. The share of deep targets separates the rows, and it is what the anchors change: 50 anchors,
one per two dozen block cells, already remove 96\% of them. The dose is not monotone at the top,
47.3\% at 200 anchors against 48.7\% at 100, within the seed spread of the latter; 200 anchors
also remove 200 targets' worth of annihilated content, which is the first condition beginning
to bind. The truncation costs about a point on the intact masks and five on the count-fixing
variant (42.48 for its RTX 5090 re-run against 37.83 truncated, on the same hardware) because the cells it removes are, in the
second case, anchors. AvgDist separates the anchored rows from the rest as the deep share does;
neither orders the unanchored rows (33.7--37.8\%), which lie within two seed standard
deviations of the intact row. The row that
hides 100 outside tokens adds a hundred targets that sit next to context and removes a hundred
context tokens: it adds no unrecoverable content and lowers the deep share only from 55\% to
51\%, so the account predicts at most a small gain; the observed drop of 2.3 points is not
significant (Welch $p = 0.09$), although no seed of the two rows overlaps, and the account does
not explain it.

\paragraph{The block families without their random fill.} At 90\% masking, four non-overlapping
blocks cannot cover 1{,}411 of 1{,}568 tokens, so the collator behind Table~\ref{tab:video}
reaches the budget with randomly placed target cells: 31\% of the targets for tube and frame
blocks and 39\% for moving blocks. Batched by padding with the blocks as drawn and the context
thinned to 157 as before, tube blocks score 35.45, 35.92 and 35.34\% and frame blocks 38.70,
40.39 and 40.97\%, against 36.17 $\pm$ 1.48 and 40.33 $\pm$ 2.19 with the fill; the ordering of
Table~\ref{tab:video} is not the fill's. Moving blocks scored 36.16\% on one seed and collapsed (probes 6.5 and 10.0\%)
on the other two, at the first and the sixty-eighth epoch; with the fill the family trained on
both seeds. We report the row as unstable rather than as a number.

One more discrepancy belongs here: the block geometry sits below random tubes (37.5\%,
Table~\ref{tab:video}), both with the public collator on the same GPUs (35.1\%) and batched intact
(36.0\%), the reverse of V-JEPA's own masking ablation at scale. Our regime is ViT-S on UCF101
for 200 epochs at batch 32;
whether the ordering reverts with a larger model or longer training is open.

\section{Probe trajectories}
\label{app:traj}

Checkpoints at epochs 100, 200 and 300, probed under the same protocol. At epoch 100 seven
of the eight families hold 20--21\,dB of pixel content; strips is the exception at 18.7. Over the
next 200 epochs the block families push that content out and the scattered families do not
move. In linear probe the scattered and block groups never cross, which counts against the reading that easy masks are a fast
start for a curriculum; we did not run a curriculum itself.

\begin{center}
\small
\begin{tabular}{lccc c ccc}
\toprule
& \multicolumn{3}{c}{Linear top-1 (\%)} && \multicolumn{3}{c}{Pixel probe (dB)} \\
\cmidrule(lr){2-4}\cmidrule(lr){6-8}
Family & ep.\,100 & ep.\,200 & ep.\,300 && ep.\,100 & ep.\,200 & ep.\,300 \\
\midrule
random     & 38.83 & 40.76 & 40.77 && 20.01 & 19.96 & 20.30 \\
dispersed  & 39.68 & 41.97 & 42.00 && 20.14 & 20.05 & 20.25 \\
strips     & 39.24 & 40.88 & 40.29 && 18.72 & 18.81 & 18.90 \\
12 blocks  & 47.29 & 49.54 & 49.14 && 20.21 & 19.15 & 18.77 \\
quadrants  & 56.99 & 58.68 & 56.95 && 21.11 & 18.43 & 17.72 \\
4 blocks   & 58.88 & 64.76 & 64.30 && 20.84 & 19.18 & 18.70 \\
1 block    & 61.45 & 65.29 & 62.92 && 20.05 & 18.27 & 17.68 \\
2 blocks   & 62.21 & 66.71 & 65.34 && 21.04 & 18.80 & 18.19 \\
\bottomrule
\end{tabular}
\end{center}

\section{Mixing easy and hard masks}
\label{app:mixing}

\begin{figure}[t]
\centering
\includegraphics[width=\linewidth]{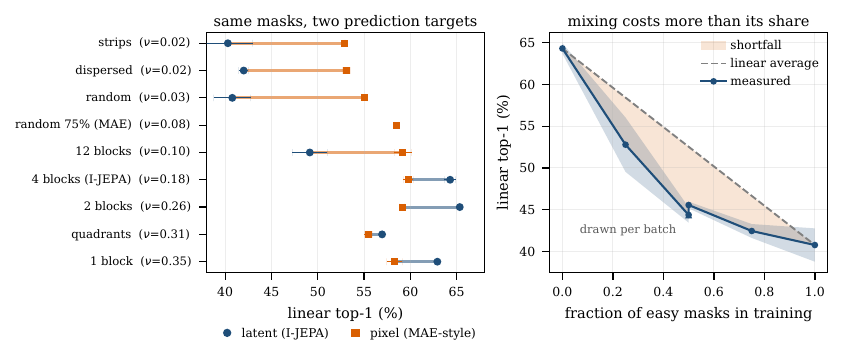}
\caption{Left: each family under the two prediction targets; a line points right when the
latent target wins. Right: every mixture of easy and hard masks lands below the linear
interpolation of the pure runs (shaded).}
\label{fig:maemix}
\end{figure}

Each image's geometry is drawn from random scatter with probability $p$ and four blocks
otherwise. Every mixture falls below the linear interpolation of the pure runs (Figure~\ref{fig:maemix}), and drawing
the geometry once per batch, which is how I-JEPA samples its block size, scores 1.2 points
below drawing it per image, a difference two seeds do not resolve. The mixtures' pixel probe sits at 19.7--20.4\,dB, the scattered level.

\begin{center}
\small
\begin{tabular}{lcccc}
\toprule
Condition & Easy fraction & Linear top-1 & Linear average & Gap \\
\midrule
pure 4 blocks $\dagger$ & 0.00 & 64.30 $\pm$0.62 & --- & --- \\
25\% easy, per image    & 0.25 & 52.78 $\pm$3.25 & 58.42 & $-5.64$ \\
50\% easy, per image    & 0.50 & 45.56 $\pm$0.40 & 52.54 & $-6.98$ \\
50\% easy, per batch    & 0.50 & 44.37 $\pm$0.92 & 52.54 & $-8.17$ \\
75\% easy, per image    & 0.75 & 42.45 $\pm$0.83 & 46.65 & $-4.20$ \\
pure random scatter     & 1.00 & 40.77 $\pm$1.99 & --- & --- \\
\bottomrule
\end{tabular}
\\[2pt]
{\footnotesize $\dagger$ I-JEPA-like geometry (F4), as in Table~\ref{tab:stage1}.}
\end{center}

A shared moving-average target encoder would explain the shortfall, by the mechanism the
frozen-target control of Appendix~\ref{app:frozen} isolates: easy samples pull its features toward
low-level content, so the hard samples are graded against low-level targets and the semantic
pressure is removed at the source. We did not train a mixture against a frozen target. The mechanism offers a reading of I-JEPA's Table~8, where the
target scale range $(0.075, 0.2)$ scores 19.2 against 54.2 at $(0.15, 0.2)$: because block
size is drawn once per batch, that setting is a per-batch mixture of small and large blocks,
the condition with the largest shortfall among our four; we have not tested that reading directly.

\section{The frozen-target control}
\label{app:frozen}

Section~\ref{sec:method} places the shortcut in the target the encoder produces: a low-level
encoder writes low-level targets, and under a recoverable mask those targets are predictable
from low-level context features, so the pair is a fixed point. A frozen target takes that
freedom away. If the geometry acts through the target, the gap between scattered and block masks
should close; if scattered masks are uninformative in themselves, it should stay.

\paragraph{Protocol.} The teacher is the target encoder of the four-block, fixed-budget run with
seed 2 on RTX 5090 (61.44\% linear, 50.88 $k$-NN) at epoch 300. Its weights are loaded into the
target encoder at initialisation and receive no moving-average update; the context encoder and
predictor start from scratch. Everything else is the recipe of Section~\ref{sec:setup}: the same
masks ($n_T = 92$, $n_C = 76$, one predictor pass), optimiser, schedule, 300 epochs and
layer-normalised targets. The student is probed at its online encoder, since its target encoder
is the teacher. We trained two seeds each of the random and four-block families against the
teacher and, on the same pool of RTX 5090 machines, two seeds of the standard random family and
one of the standard four-block family as references. The reference lands at 60.16\%, inside the
60--63 window we had set for reading the two machine pools together and 0.8 below the lowest of the three
earlier RTX 5090 seeds (60.98, 61.44 and 62.20).

\paragraph{Prediction fixed before training.} The account predicts the gap to fall from about
twenty points to single digits; the reading that block masks are needed in themselves predicts it
to stay near twenty. The decision rule was a gap of at most 8 points in favour of the account and
of at least 15 against it, anything between left open, with no prediction for either student's
absolute level.

\begin{center}
\small
\begin{tabular}{llcccc}
\toprule
Target & Family & Seeds & Linear top-1 & $k$-NN & Pixel probe (dB) \\
\midrule
moving average & random   & 2 & 42.08 (41.64, 42.52) & 30.61 & 20.19 \\
moving average & 4 blocks & 4 & 61.20 (60.16--62.20)  & 51.42 & 18.33 \\
frozen teacher & random   & 2 & 67.32 (67.20, 67.44) & 52.38 & 21.05 \\
frozen teacher & 4 blocks & 2 & 68.82 (68.30, 69.34) & 54.88 & 19.91 \\
\bottomrule
\end{tabular}
\end{center}

\paragraph{Result.} With the moving target the two families differ by 19.1 points (Welch
$t = 31$, $p = 10^{-4}$); against the frozen teacher they differ by 1.5 ($t = 2.8$ on about one
degree of freedom, $p = 0.20$), below the threshold the rule set for the account. Three
details bear on the reading. Both students pass their teacher, 67 to 69 against 61.4, as SALT finds for students of
a frozen pixel-trained teacher \citep{li2025salt}, so the control is not a ceiling in which every
mask approaches the teacher. The pixel probe keeps the direction of Section~\ref{sec:levels}:
the random student holds 21.1\,dB of pixel content and the block student 19.9, against 20.2 and
18.3 with the moving target, so the geometry still decides how much low-level content the encoder
carries while the target decides its level. And the random student has the easier task, a final
loss of 0.116 against 0.161 for blocks, yet a representation as semantic, so the difficulty of the
task does not set the level the encoder reaches.

\paragraph{What the control settles.} It settles where the geometry acts: through the target the
encoder produces. With the target fixed, what the encoder must represent is set by the target,
and the teacher's features are semantic under either mask. With two seeds per family the Welch
95\% interval of the gap runs from $-3.9$ to $6.9$ points on 1.1 degrees of freedom: its upper end
falls below the 8-point line of the rule, although that width rests on the spread of the two block
seeds, and a gap of a few points is not excluded. The teacher was itself trained with four blocks; a teacher
trained with scattered masks, whose features are low-level, is the natural next control, and the
account predicts that the gap would partly return, since low-level targets are again recoverable
from the context under a scattered mask. The absolute levels of the frozen and moving-target rows
are not comparable, because the teacher's own 300 epochs precede the student's; the comparison
that carries the test is the gap between the two families under each target. The frozen-target students are probed at the
online encoder and the moving-target models at the target encoder, each the encoder its recipe
trains for downstream use. All runs in the table share one software stack and one kind of GPU
(Appendix~\ref{app:impl}).

\section{Probe views for the whole-frame families}
\label{app:views}

A model trained on a thin context slab sees a different input at probe time, so the poor score
of whole-frame masks might be a distribution shift. We probed the eight seed-0 checkpoints twice
more, feeding the frozen encoder only its own training context masks and then only a fixed
middle tubelet. The gap between the six block and scatter families and the two families that
remove whole time steps is 6.65, 6.57 and 6.51 points across the three views (Table~\ref{tab:video}'s two-seed means give 6.1
for the standard view), so the artefact is ruled out.

\section{Space-time atoms and the video families}
\label{app:video3d}

On video an atom has two extents, a temporal one and a spatial one. We measure the detail
energy of 64 real clips over pairs of extents with a separable Haar analysis, 16 frames at
$224\times224$. Rows are the temporal extent, columns the spatial extent, entries are
percentages of total detail energy; the first column is temporal detail with no spatial
detail, and the entries sum to 100 up to rounding.

\begin{center}
\small
\begin{tabular}{lcccccccc}
\toprule
Temporal extent & none & 2\,px & 4\,px & 8\,px & 16\,px & 32\,px & 64\,px & 128\,px \\
\midrule
static (no temporal detail) & --- & 1.0 & 1.7 & 3.1 & 5.9 & 10.8 & 12.6 & 7.3 \\
2 frames                    & 0.6 & 1.3 & 1.9 & 2.8 & 3.8 & 4.6 & 3.2 & 1.1 \\
4 frames                    & 1.1 & 0.7 & 1.1 & 1.7 & 2.5 & 3.4 & 2.9 & 0.9 \\
8 frames                    & 1.9 & 0.4 & 0.7 & 1.1 & 1.8 & 2.7 & 2.7 & 1.5 \\
16 frames                   & 2.1 & 0.2 & 0.4 & 0.8 & 1.4 & 2.3 & 2.3 & 1.5 \\
\bottomrule
\end{tabular}
\end{center}

Content with temporal detail carries 57.5\% of the energy, concentrated at 2--4 frames and
16--64 pixels, and static content carries 42.5\%. A mask that leaves a neighbouring tubelet
visible at the same spatial location therefore leaves most of the moving content recoverable,
which is why the recovery baseline has to include temporal copying and why $\nuc^{3d}$ alone
misranks whole-frame removal at the lower masking ratio.

\paragraph{Families.} On the $T \times G \times G$ tubelet grid: \emph{tube, $k$ blocks}
places $k$ spatial blocks and extrudes them over all tubelets; \emph{frame blocks} places the
same block statistics independently in every tubelet; \emph{moving blocks} drifts each block
by up to one patch per tubelet; \emph{random tube} repeats one random patch set in every
tubelet; \emph{random tokens} scatters over the whole grid; \emph{whole frames} removes entire
tubelets, choosing non-adjacent ones while the budget allows; \emph{half of time} removes the
last tubelets. Context is the complement, subsampled to a fixed count.

\section{Audio: a modality where the account does not hold}
\label{app:audio}

We repeated the analysis on log-mel spectrograms, pre-training the same architecture on the
AudioSet balanced set \citep{gemmeke2017audioset} (18{,}683 clips, $1024\times128$ inputs, patch 16, 75\% masked, 300
epochs, one seed) and evaluating by a five-fold linear probe on ESC-50 \citep{piczak2015esc50} and a linear mAP probe
on AudioSet.

\begin{center}
\small
\begin{tabular}{lcccc}
\toprule
Family & $\nuc$ (anisotropic) & Null-space share & ESC-50 linear & AudioSet mAP \\
\midrule
time strips + frequency band & 0.126 & 0.136 & 67.10 & 10.50 \\
random tokens                & 0.020 & 0.062 & 64.25 & 9.71 \\
8 thin time strips           & 0.058 & 0.087 & 60.95 & 9.44 \\
3 time strips $\ddagger$     & 0.277 & 0.252 & 59.15 & 9.50 \\
1 block                      & 0.185 & 0.126 & 55.30 & 8.96 \\
1 wide time strip            & 0.459 & 0.210 & 53.80 & 8.15 \\
4 blocks $\dagger$           & 0.073 & 0.090 & 52.55 & 8.86 \\
2 frequency bands            & 0.054 & 0.091 & 45.20 & 4.42 \\
\bottomrule
\end{tabular}
\end{center}

{\footnotesize $\dagger$ and $\ddagger$: the random-block and time-frequency families of
\citet{fei2023ajepa}.}

Two observations. The published ordering reproduces: A-JEPA's time-frequency family (3 time strips,
$\ddagger$) beats its random-block family (4 blocks, $\dagger$) by 6.6 points, and the difference is positive in all five folds. But the reason we would have
given is wrong. The null-space share does not order the families ($\rho = -0.10$), and the
family with the least annihilated content is second best. The measurement view explains why:
spectrogram energy is extremely anisotropic, with 70.7\% of the coarse detail energy on atoms
that are long in time and span all mel bins, and no mask family opens a recoverability gap
wider than 1.4\,dB once the low-level prior is allowed to copy along time. Every family leaves
nearly the same content recoverable, so the first condition has little to separate, and the
quantity that orders the families on images and video does not order them here. What remains is shape: the family whose targets are adjacent to context along
both axes is best, and the family whose context is two narrow frequency bands is worst by far, which
we cannot cleanly attribute to reachability because the probe sees a full spectrogram while
the encoder was trained on a slab. We report the modality as a boundary of the account rather
than as a result.

\section{Generated images}
\label{app:synthetic}

On 96 images generated by Stable Diffusion 1.4 \citep{rombach2022ldm} for three ImageNet classes, compared with 96 real validation images on the $14\times14$ grid of the main text (Appendix~\ref{app:realenergy} uses $16\times16$), the family ordering of TV inpainting PSNR is preserved at
$\rho = +0.976$, null-space energy is higher for every family (for example 28.1\% against
24.5\% for one block), PSNR is 0.4--1.3\,dB lower throughout, and the scattered-versus-block
gap is 4.58\,dB against 4.28\,dB. The detail-energy spectrum carries less 128\,px content
(26.2\% against 32.0\% for these 96 real images; 32.3\% over the 512 of Appendix~\ref{app:derivation}) and more 16--32\,px content, and mean pixel gradients are nearly equal (0.0240 against 0.0236). The recoverability ordering behind the first condition transfers to generated images. We did not train
on them, and generator fingerprints are a low-level regularity that $\nuc$ does not model.

\section{Implementation}
\label{app:impl}

\paragraph{Data.} ImageNet-100 is the 100-class subset of ImageNet \citep{deng2009imagenet}
introduced by \citet{tian2020cmc}, with 126{,}689 training and 5{,}000 validation images; we
use the copy at \nolinkurl{huggingface.co/datasets/clane9/imagenet-100}, a third-party mirror,
under ImageNet's terms of access for non-commercial research. UCF101, AudioSet and ESC-50 are
used under their own research terms; the generated images of Appendix~\ref{app:synthetic} come
from Stable Diffusion 1.4 under its CreativeML OpenRAIL-M licence.

\paragraph{Masks.} Images are $224\times224$ with $16\times16$ patches, a $14\times14$ grid.
Target families draw an exact patch count; the context is I-JEPA's block minus the targets,
subsampled to a fixed count so that every family in a matched set sees the same number of
patches. Random draws the target patches uniformly; dispersed draws them by farthest-point
sampling on the grid; each block family places $k$ non-overlapping rectangles of equal area with
aspect ratios drawn from $(0.75, 1.5)$, I-JEPA's range, and adjusts the count at the boundary of
the last block; rasterised quadrants hide whole $7\times7$ quadrants in random order until the
budget is met, adjusting at the last one; the selected placements of Section~\ref{sec:causal} draw
16 four-block candidates per image and keep the one with the highest or the lowest $\nuc$ against
a provisional context block drawn as I-JEPA draws its own. Strips choose rows with no two adjacent while the budget allows; above 98 target
patches adjacency is forced, which is why the family crosses over at large budgets
(the ordering of the eight families holds from 20\% to 61\% masking).

\paragraph{Recovery.} Harmonic inpainting uses coarse-to-fine Jacobi relaxation. TV \citep{rudin1992tv} uses
Chambolle--Pock \citep{chambolle2011firstorder} with 1{,}000 iterations, $\tau=0.2$, $\sigma=0.5$, initialised at the harmonic
solution. Wavelet-$\ell_1$ uses iterative soft thresholding with a geometrically decreasing
threshold under the data constraint, db4 \citep{daubechies1988wavelets} with 5 levels.

\paragraph{Training.} ViT-S/16, predictor depth 6 and width 384, batch 256, 300 epochs,
AdamW, learning rate $3\times10^{-4}$ with 30 warmup epochs, weight decay 0.04 to 0.4, EMA
0.996 to 1.0, bfloat16. The pixel-target ablation uses MAE's learning rate $1.5\times10^{-4}$
and weight decay 0.05, with 20 warmup epochs (MAE uses 40 over a longer schedule); its models are probed at the online
encoder, the only encoder they have. The frozen-target runs of Appendix~\ref{app:frozen} load the
teacher's weights into the target encoder and skip its moving-average update, with everything
else unchanged; they too are probed at the online encoder.

\paragraph{Predictor passes.} The fixed-budget families run the predictor once over the union of
the target patches; the official recipe runs it once per target block. On the native sampler
the two differ by 1.3 points (59.55 against 58.23, Appendix~\ref{app:native}).

\paragraph{Evaluation.} Linear probe: SGD with momentum on frozen average-pooled tokens, 100
epochs, cosine schedule, learning rate swept over $\{0.8, 0.3, 0.1, 0.05, 0.01, 0.002\}$, best
reported. The optimum sits at the top of the grid, 0.8, in 55 of 94 image runs (pixel-target runs included) and 15 of 57 video runs; in those runs the curve is flat at the top: the difference between 0.8 and 0.3 is at most 0.22 points on images and 0.48 on video. The learning rate is selected on the same validation split that reports the score; the grid is identical for every run and the flat top makes the choice immaterial. $k$-NN uses $k=20$ with cosine similarity and
temperature 0.07. The pixel probe regresses each frozen token onto its $8\times8\times3$
downsampled patch on one million training tokens and reports PSNR on all validation tokens.

\paragraph{Video training.} ViT-S with a $2\times8\times8$ tubelet embedding on
$16\times112\times112$ clips, giving 1568 tokens; predictor depth 6 and width 384 with a
learned mask token; L1 loss to layer-normalised moving-average targets; batch 32, 200 epochs,
AdamW, learning rate $2\times10^{-4}$ with 20 warmup epochs, weight decay 0.04 to 0.4, moving
average 0.996 to 1.0 (I-JEPA's value; V-JEPA's release uses 0.998), bfloat16. Clips are 16 frames at stride 4 with a random resized crop and
horizontal flip shared across frames. The probe uses three evenly spaced clips per video with
a short-side resize and centre crop, averaged before the linear layer.

The 46 scored image checkpoints of the sweep are 22 configurations at two seeds, with four runs
for the native sampler because it was trained with one and with four predictor passes.

\paragraph{Hardware.} The sweep, the pixel-target and mixing runs, the video families and the collator rows marked $\ast$ in
Table~\ref{tab:reach} (the public collator, and the count-fixing variant with and without thinned context)
were trained on H100 and H200 GPUs. The other collator rows, including a re-run of the count-fixing
variant and its truncated version, the block families without fill, the pool test, three of the five selection seed pairs and three seeds of the
unselected four-block draw were trained on RTX 5090 GPUs with the same code,
configurations and data, verified file by file; video runs there use batch 16 with two
gradient-accumulation steps, which the schedules treat as one step of 32. On video the two
agree on the one configuration run on both (the count-fixing variant, one RTX 5090 seed at
42.48\% against 42.76 $\pm$ 1.29 on H100);
on images the RTX 5090 runs score lower for the same configuration, by 2.8 points for four
blocks at fixed budget (F4: 61.54 $\pm$ 0.62 against 64.30 $\pm$ 0.62), by 0.7 for the
highest-$\nuc$ selection (64.03 $\pm$ 1.17 against 64.70 $\pm$ 0.08) and by 4.1 for the lowest
(56.15 $\pm$ 2.01 against 60.22 $\pm$ 1.84), and higher by 1.3 for random scatter (42.08 on the
second RTX 5090 pool against 40.77), within that family's seed spread; every image comparison
in the paper is therefore made within one kind of hardware. On video that single-seed agreement is all we have, and Table~\ref{tab:fix} compares the public-collator row (H100/H200)
with rows trained on RTX 5090. Each run's hardware is recorded by the directory that holds its evaluation file (\texttt{results/runs\_5090/} and \texttt{results/runs\_5090b/} for RTX 5090). The
cluster runs used PyTorch 2.5.1 with CUDA 12.4 on Python 3.11, the RTX 5090 runs PyTorch 2.11.0
with CUDA 12.8 on Python 3.12; we have not traced the image offset to a component. We treat the cluster's H100 and H200 GPUs as one kind
of hardware and write H100 for both. The seven
frozen-target and reference runs of Appendix~\ref{app:frozen} ran on a second pool of RTX 5090
machines with the same software stack; its four-block reference scores 60.16\% against
$61.54 \pm 0.62$ on the first.

\paragraph{Collapse.} Every latent-target run logs the effective rank of the layer-normalised
target features on 4{,}096 tokens (the exponential of the entropy of the normalised singular
values) every 20 iterations; one seed of the intact padded video collator recorded no value. Healthy runs end between 190 and 350; three runs fell below 20 and stayed
there: moving blocks without fill, seed 0 at epoch 1 and seed 2 at epoch 68, and the low-$\nuc$
pool, seed 2 at epoch 6. Their probes (6.5, 10.0 and 10.1\%) are excluded from every mean and
reported where the run appears.

\paragraph{Compute.} 102 image and video pre-training runs on H100 and H200 GPUs, plus 8
audio runs on the same cluster (Appendix~\ref{app:audio}): 72 image runs of 300 epochs at 56--62 seconds per epoch
(about 350 GPU-hours) and 30 video runs of 200 epochs at 55--75 seconds per epoch (about 110
GPU-hours), one GPU per run, plus roughly 40 GPU-hours for the probes and the training-free scoring. A further 49 runs on
RTX 5090 GPUs, 22 image runs at 72--110 seconds per epoch and 27 video runs at 100--150
seconds per epoch, took about 365 GPU-hours.

\section{Scope of the evidence}
\label{app:scope}

This appendix states what each part of the evidence supports and the choices a reader would
want to know about before weighing it.

\paragraph{What the account adds to the inscribed square and AvgDist.} Across the sweep the three
quantities rank the configurations alike (Appendix~\ref{app:avgdist}), and the account does not
rest on $\nuc$ ranking better. Its contribution is the reading: all three measure how much of the
target no low-level prior can recover, which is why one reading accounts for the recovery algorithms
of Table~\ref{tab:stage1}, the level change of Section~\ref{sec:levels}, the behaviour of strips
and whole frames, and the response to a pixel target and to a frozen one. $\nuc$ is the computable
form of that reading. The pool test reaches where the sweep does not: it holds the deepest square
and the mean distance fixed, and the multi-scale amount still separates the two pools by 7.4 points.

\paragraph{Two conditions on one axis.} Over the 22 configurations $\rho(\nuc, \text{reach}) =
-0.995$, so the sweep itself shows a rise and a fall along one axis, the shape of SimMIM's AvgDist
ridge \citep{xie2022simmim} and of InfoMin's sweet spot \citep{tian2020infomin}. The two
failure sides are told apart on video: at 50\% masking recoverability under temporal copying orders
the families, and at 90\%, with every family past the first condition, the share of deep targets
does. On images the pool test separates the amount from the deepest square and the mean distance
but not from reachability. The thresholds of Section~\ref{sec:plateau} are read off the runs they summarise and
describe a graded response; both pools clear them and still differ by 7.4 points. The reachability
condition rests on three configurations, one of them the native sampler, whose context is also
truncated (Appendix~\ref{app:native}).

\paragraph{One direction of the argument.} The argument of Section~\ref{sec:method} shows that a recoverable
target makes the low-level map near-optimal and, with a moving target, self-consistent. It does
not say which solution training selects when the target is unrecoverable: representing content is
one, collapse is another,
and what holds it off is the moving average and the predictor asymmetry the recipe inherits from
BYOL \citep{grill2020byol}. The effective-rank logs show that the runs which reach the semantic
level do not collapse, and the three that did are excluded from every mean (Appendix~\ref{app:impl}); the
frozen-target control shows where the geometry acts. The selection of content over collapse is
inherited from the recipe and measured, not derived.

\paragraph{Choices made after seeing the runs.} The two thresholds of Section~\ref{sec:plateau}, the four-versus-four
boundary at $\nuc = 0.1$ and the video statistic were chosen after the runs; on video, $\nuc^{3d}$
does not order the 90\% families and the share of deep targets does, and Table~\ref{tab:video}
reports $\nuc^{3d}$, the recovery PSNR and that share, with the context density in its footnote.
Two tests were written down before training, with their decision rules: the pool test
(Appendix~\ref{app:avgdist}) and the frozen-target control (Appendix~\ref{app:frozen}). Both came
out on the account's side, the pool by more than predicted. A feature-space score built on a frozen
DINO encoder reproduces the scattered-versus-block split but does not order the block families, so
reachability is measured on the mask instead (Appendix~\ref{app:feature}).

\paragraph{Seeds and tests.} Most configurations have two seeds, and the claims rest on different
amounts of evidence. The family split rests on a 19.3-point gap, with every seed of the four
high-$\nuc$ families above every seed of the four low-$\nuc$ ones. The selection effect rests on five paired seeds ($t = 6.3$). The anchors rest on three
seeds against three (Welch $p = 3 \times 10^{-4}$); hiding a hundred tokens outside the blocks uses
the same design ($p = 0.09$). The pool test rests on three seeds against two with no overlap
($p = 0.17$ on about one degree of freedom). The intact-against-truncated comparison uses runs
from two kinds of hardware ($p = 0.36$); the thinning control, run on one kind, supports the reading
that the truncation costs little.
The frozen-target gap is 1.5 points with $p = 0.20$. The differences between the Spearman
coefficients over 22 configurations carry bootstrap intervals (Appendix~\ref{app:avgdist}); those over eight video families are
ordering statements, with no correction for the several statistics tried.

\paragraph{Hardware and probes.} Image runs on H100 and on RTX 5090 differ by 0.7 to 4.1 points
for the same configuration, in a direction that depends on the configuration, so every image
comparison is made within one kind of hardware; the video cross-check rests on one configuration
and one seed (Appendix~\ref{app:impl}). The linear probe's learning rate is chosen on the
validation split that reports the score, from a grid identical for every run, whose top is flat
to within 0.22 points on images.

\paragraph{What the frozen-target runs are.} They are distillation from a frozen teacher, the
setting SALT studies with a pixel-trained teacher \citep{li2025salt}, and that both students pass
their teacher reproduces SALT's finding. We use the setting as a control on the mechanism.

\paragraph{Where the account stops.} It is established on ViT-S with ImageNet-100 and UCF101. On
video, V-JEPA's intact geometry scores below random tubes in our setting, the reverse of V-JEPA's
own ablation (Appendix~\ref{app:native}), so the anchor result is established in that regime. Under
a pixel target at 75\% masking, MAE's Table 1(f) shows block masks losing to random ones, so how
much geometry matters to a fixed target depends on the ratio and the probe. On spectrograms the
account fails (Appendix~\ref{app:audio}).

\end{document}